\documentclass{article}

\usepackage{graphicx}
\usepackage{amsmath}
\usepackage{bm} 
\usepackage{subcaption}
\usepackage{float} 
\usepackage{multirow}
\usepackage{tabularx}
\usepackage[table,xcdraw]{xcolor}
\usepackage{wrapfig}
\usepackage{caption}
\usepackage{enumitem}

\usepackage[final, nonatbib]{neurips_2023}

\usepackage[numbers, sort&compress]{natbib}

\usepackage[utf8]{inputenc} 
\usepackage[T1]{fontenc}    
\usepackage{hyperref}       
\usepackage{url}            
\usepackage{booktabs}       
\usepackage{amsfonts}       
\usepackage{nicefrac}       
\usepackage{microtype}      
\usepackage{xcolor}         
\usepackage{titlesec}

\title{RE-tune: Incremental Fine Tuning \\of Biomedical Vision-Language Models \\for Multi-label Chest X-ray Classification
}

\author{
  Marco Mistretta, Andrew D. Bagdanov\\
  Università degli Studi di Firenze\\
  \texttt{marco.mistretta@unifi.it, andrew.bagdanov@unifi.it} \\
}

\newcommand{\minisection}[1]{\vspace{0.0in} \noindent {\bf #1}}

\begin{document}
\setlength{\parskip}{5pt}
\maketitle

\begin{abstract}
  In this paper we introduce \textbf{RE-tune}, a novel approach for fine-tuning pre-trained Multimodal Biomedical Vision-Language models (VLMs) in Incremental Learning scenarios for multi-label chest disease diagnosis. RE-tune freezes the backbones and only trains simple \emph{adaptors} on top of the Image and Text encoders of the VLM. By engineering positive and negative text prompts for diseases, we leverage the ability of Large Language Models to steer the training trajectory. We evaluate RE-tune in three realistic incremental learning scenarios: class-incremental, label-incremental, and data-incremental. Our results demonstrate that Biomedical VLMs are natural continual learners and prevent \textit{catastrophic forgetting}. RE-tune not only achieves accurate multi-label classification results, but also prioritizes patient privacy and it distinguishes itself through exceptional computational efficiency, rendering it highly suitable for broad adoption in real-world healthcare settings.
\end{abstract}

\section{Introduction}
\label{sec:1}
Chest X-rays are a cornerstone of non-invasive disease diagnosis, but their interpretation poses unique challenges and relies on discernment of subtle gray-scale variations. Misinterpretation can lead to delayed treatment or missed diagnoses, emphasizing the need for more efficient and accurate diagnostic methods. The advent of deep vision models has yielded remarkable advances in medical image analysis~\citep{irvin2019chexpert}. Simultaneously, transformer-based Large Language Models (LLMs) have empowered AI systems to comprehend the complex language of radiology reports~\citep{vaswani2023attention, devlin2019bert}.

Synergy between vision and language has led to the emergence of Multimodal Biomedical Vision-Language models (VLMs) capable of interpreting both images and associated reports. Biomedical VLMs offer a more comprehensive analysis of patient condition~\citep{zhang2022contrastive, 9710099, boecking2022making}. Despite promising capabilities, significant challenges persist in the practical implementation of these technologies. When zero-shot performance is insufficient, VLMs require massive computational power and massive datasets for fine-tuning, posing feasibility issues in resource-limited environments. Furthermore, medical data are constantly generated in hospitals, and traditional models are susceptible to \textit{catastrophic forgetting}, where the introduction of new data leads to the loss of previously learned information.

To overcome these challenges, incremental learning techniques have been developed~\citep{masana2022classincremental}, but many rely on storing of training data (exemplars), which raises important privacy and security concerns particularly relevant in the context of healthcare. In this paper we propose RE-tune, a privacy-preserving, computationally efficient, and exemplar-free approach to incremental fine-tuning of Biomedical VLMs for multi-label chest disease classification.
\section{Proposed Method}
\label{sec:2}
\begin{figure}
    \begin{subfigure}{0.49\textwidth}
        \centering
        \includegraphics[width=0.7\linewidth]{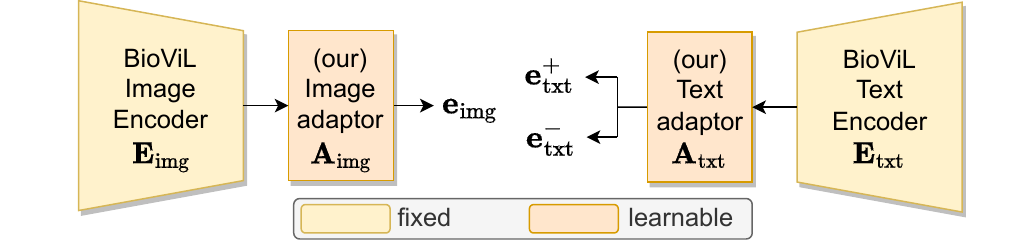} 
        \caption{Freeze and add \textit{adaptors}.}
        \label{fig:freeze}

        \tiny
        \begin{tabularx}{0.7\linewidth}{l|X}
            \hline
            \multirow{2}{*}{\textbf{\begin{tabular}[l]{@{}l@{}}Template\\ Prompts\end{tabular}}}    &\hspace{-0.19cm}$+$) There is evidence of \_\_\_\_\_\_             \\ \cline{2-2} 
            &\hspace{-0.19cm}$-$) No findings of \_\_\_\_\_\_                \\ \hline
            
            \multirow{2}{*}{\textbf{\begin{tabular}[l]{@{}l@{}}Generative\\ Prompts\end{tabular}}} &\hspace{-0.19cm}$+$) Small residual right pleural eff.             \\ \cline{2-2} 
            &\hspace{-0.19cm}$-$) No evidence of injury in the chest \\ \hline
            
            \multirow{2}{*}{\textbf{\begin{tabular}[l]{@{}l@{}}Random\\ Prompts\end{tabular}}} &\hspace{-0.19cm}$+$) \textit{some random words}             \\ \cline{2-2} 
            &\hspace{-0.19cm}$-$) \textit{some random words} \\ \hline
        \end{tabularx}
        \caption{Define Text Prompts.}
        \label{tab:prompts}
        
        \includegraphics[width=0.7\linewidth]{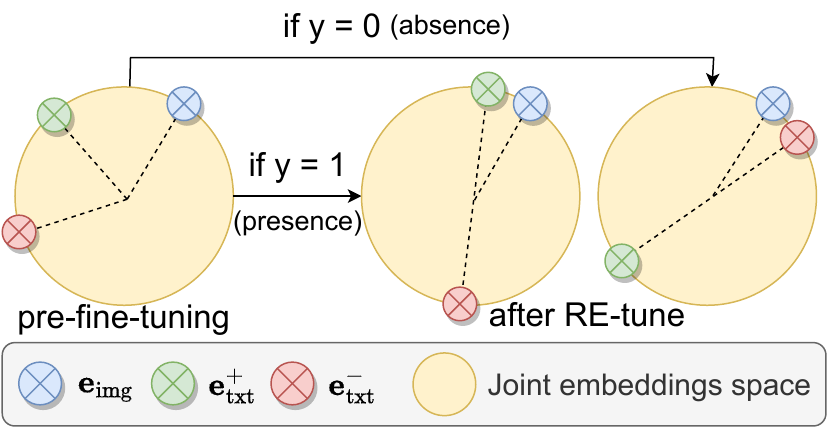}           
        \caption{Fine-tune \textit{adaptors} (RE-tune the model).}
        \label{fig:cosine}
        
    \end{subfigure}
    \begin{subfigure}{0.49\textwidth}
         \centering
        \includegraphics[width=0.9\linewidth]{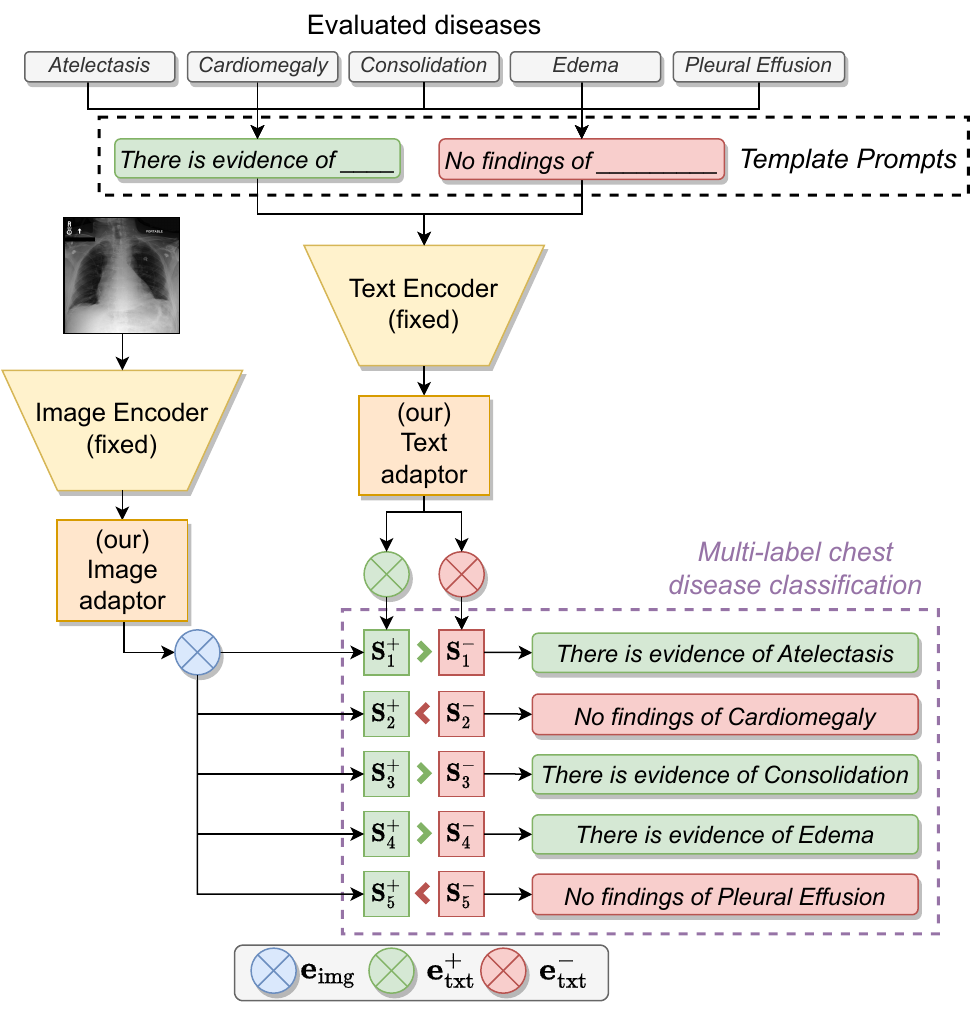}
        \caption{Inference procedure.}
        \label{fig:inference}

    \end{subfigure}
    \caption{RE-tune fine-tuning steps (left) and inference procedure (right).}
\end{figure}
Our experiments focus on fine-tuning BioViL~\citep{boecking2022making}, a publicly available state-of-the-art Multimodal Biomedical model. BioViL's zero-shot performance and its specialized text modeling makes it the ideal candidate to fine-tune. Importantly, BioViL has \textit{not been pre-trained on CheXpert}~\citep{irvin2019chexpert}, 
a well-established and widely adopted dataset of multi-labelled chest X-rays, we use in our experiments. 

\minisection{Freezing backbones and extending with \textit{adaptors}.} The initial phase involves the straightforward step of freezing the backbones: the Image Encoder $\mathbb{\textbf{E}}_{\text{img}}$ and the \mbox{Text Encoder $\mathbb{\textbf{E}}_{\text{txt}}$} (Fig.~\ref{fig:freeze}).
This ensures that the learned representations at their core remain intact during fine-tuning and reduces computational demands.
Afterward we expand the pre-trained architecture by adding additional components, that we call \textit{adaptors} , on top of the Encoders (Fig.~\ref{fig:freeze}). Simple \textit{adaptors} such as Dense Layers or Multi Layers Perceptrons have been evaluated in all the combinations, including applied exclusively to the Image Encoder, exclusively to the Text Encoder, to both or shared between them. Note that the \textit{adaptors} output size is equal to its input size, which is equal to the joint embedding size. \textit{RE-tune} incrementally adapts the embedding space to \textit{REmember} previously learned information.

\minisection{Defining the text prompts.}
Consider a \textit{batch} $B$ of $N$ radiology images and corresponding labels, expressed as \mbox{$B = (x_{i}, y_{i})_{i=1}^{N}$}. Here, $x_i$ represents the input image, and \mbox{$y_i \in \{0, 1\}^C$} denotes the binary labels associated with the image, indicating the absence ($y_{i} = 0$) or presence ($y_{i} = 1$) of \mbox{$C$ \textit{evaluated different diseases}} (i.e. the \textit{five CheXpert competition tasks} defined in~\citep{irvin2019chexpert}).
For each evaluated disease, we construct \textit{a positive and a negative text prompt} to signify the presence or absence of the respective disease (Fig.~\ref{tab:prompts}). We evaluated three types of prompts: \textit{Template Prompts}, which represents the classic prompts used in zero-shot classification~\cite{boecking2022making}, \textit{Generative Prompts} that we obtain by summarizing labeled medical reports~\cite{johnson2019mimiccxrjpg, smit2020chexbert} with a recent T5 model~\cite{chizhikova-etal-2023-sinai}, and \textit{Random Prompts} that are used in order to verify if text semantics matter during fine-tuning.

\minisection{Fine-tuning.}
The Image Encoder $\mathbb{\textbf{E}}_{\text{img}}$ extracts image embeddings $\tilde{\mathbf{e}}_{\text{img}}$, and the \textit{Image Adaptor} $\textbf{A}_{\text{img}}$ further projects these embeddings ($\mathbf{e}_{\text{img}} = \textbf{A}_{\text{img}}(\tilde{\mathbf{e}}_{\text{img}})$). The positive and negative text prompts are given to the Text Encoder, which extracts the positive $\tilde{\mathbf{e}}_{\text{txt}}^{+}$ and the negative $\tilde{\mathbf{e}}_{\text{txt}}^{-}$ text embeddings. The \textit{Text Adaptor} $\textbf{A}_{\text{txt}}$ further projects these embeddings, resulting in $\mathbf{e}_{\text{txt}}^{+} = \textbf{A}_{\text{txt}}(\tilde{\mathbf{e}}_{\text{txt}}^{+})$ and $\mathbf{e}_{\text{txt}}^{-} = \textbf{A}_{\text{txt}}(\tilde{\mathbf{e}}_{\text{txt}}^{-})$.

Then, the \textit{positive cosine similarity} $\textbf{S}^{+}$, and the \textit{negative cosine similarity} $\textbf{S}^{-}$, are calculated for each label $j$ using the following formulations: 
$\textbf{S}^{+}_{j}=\frac{\mathbf{e}_{\text{img}} \cdot \mathbf{e}_{{\text{txt}}_{j}}^{\text{+}}}{|\mathbf{e}_{\text{img}}| \cdot |\mathbf{e}_{{\text{txt}}_{j}}^{\text{+}}|}, 
\textbf{S}^{-}_{j}=\frac{\mathbf{e}_{\text{img}} \cdot \mathbf{e}_{{\text{txt}}_{j}}^{\text{-}}}{|\mathbf{e}_{\text{img}}| \cdot |\mathbf{e}_{{\text{txt}}_{j}}^{\text{-}}|}\:\forall j \in [1, \dots, C].$

The loss function employed is the Binary Cross Entropy where we propose to employ the difference between the positive and negative cosine similarities as logits:
\begin{gather}
    L_{\text{BCE}} = -\frac{1}{N} \sum_{i=1}^{N} \left(\sum_{j=1}^{C} y_{ij} \log(\sigma(\textbf{S}^{+}_{ij} - \textbf{S}^{-}_{ij})) + (1 - y_{ij}) \log(1 - \sigma(\textbf{S}^{+}_{ij} - \textbf{S}^{-}_{ij}))\right).
\end{gather}
For each evaluated disease, this loss aims to drive the positive and negative text embeddings in opposite directions. As depicted in Fig.~\ref{fig:cosine}.
At inference, for each of the $C$ evaluated disease we predict the presence of diseases $j$ if $\textbf{S}^{+}_{j} \geq \textbf{S}^{-}_{j}$, otherwise we predict its absence (see Fig.~\ref{fig:inference}).

\minisection{Training and Evaluation.}
We evaluate our approach in zero-shot, joint training, and incremental scenarios. 
Zero-shot (i.e. before fine-tuning) performance is a lower bound baseline, while joint training is an upper bound since all images and labels are available in a single training session. We consider three realistic incremental scenarios: class-, label- and data-incremental.
In class- and label-incremental scenarios the number of training Tasks $T$ is equal to the number of diseases $C$. In class-incremental scenarios in each task the model is exposed to only one disease at a time. In label-incremental scenarios in each task the model is exposed to labels of the current task's disease as well as all labels from previous tasks. In data-incremental scenarios the training set is equally partitioned into $M=20$ subsets and all labels are available. Note that in all scenarios we assume the task data are disjoint and we cannot access previously-seen data due to privacy concerns. At the end of each task the average AUC is evaluated on the \textit{entire test set} consisting of previously seen and unseen diseases. For both joint and incremental scenarios we train with the same settings: Adam optimizer with learning rate of $1e-4$ for $10$ epochs.
Experiments were conducted with multiple seeds and performance averaged to assess robustness and account for random variability.

\section{Results and Discussion}
\label{sec:3}
\begin{figure}
    \begin{subfigure}{0.33\textwidth}
        \centering
        \includegraphics[width=\linewidth]{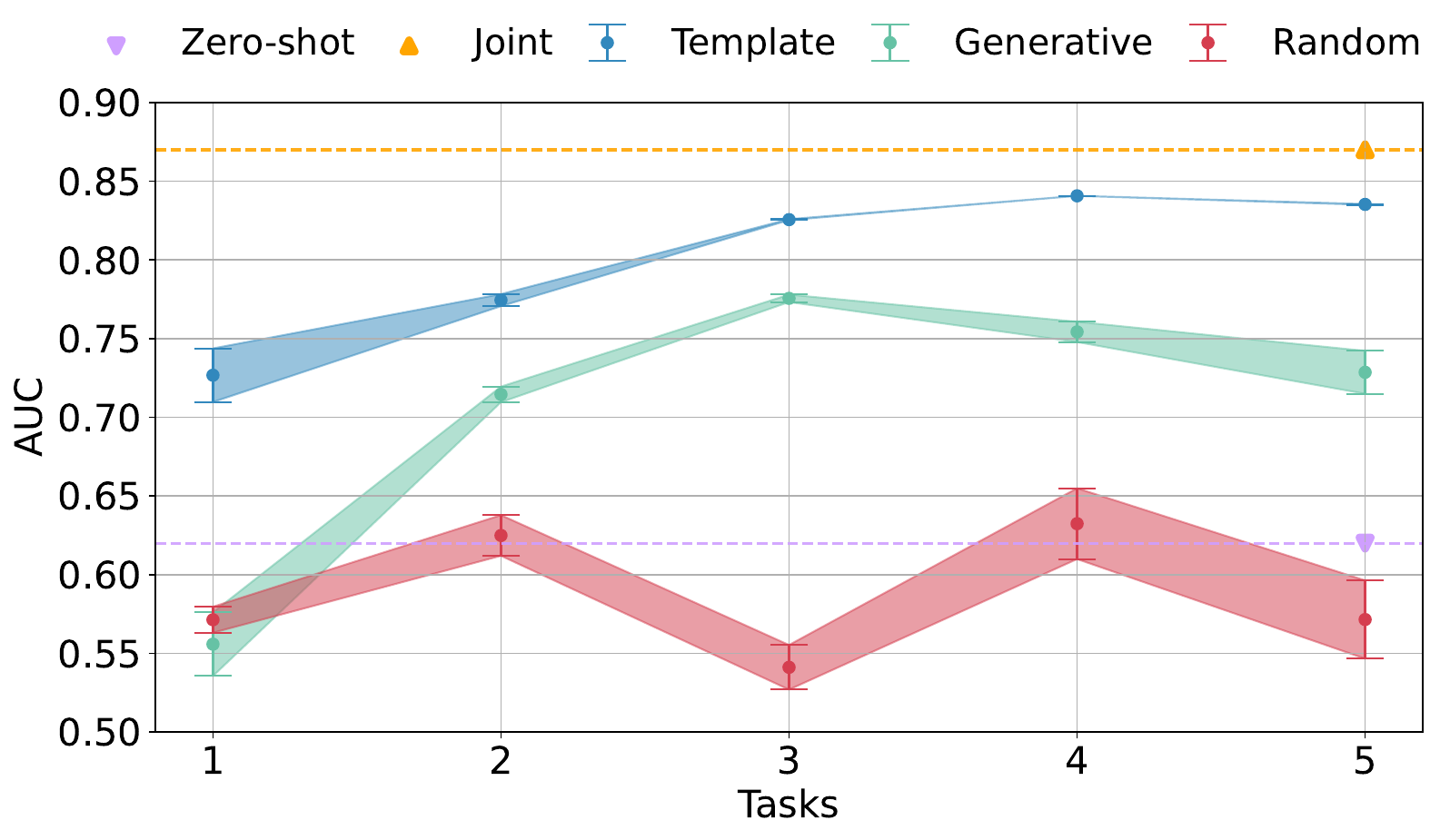}  
        \caption{Class incremental}
        \label{fig:class-sub}
    \end{subfigure}
    \begin{subfigure}{0.33\textwidth}
        \centering
        \includegraphics[width=\linewidth]{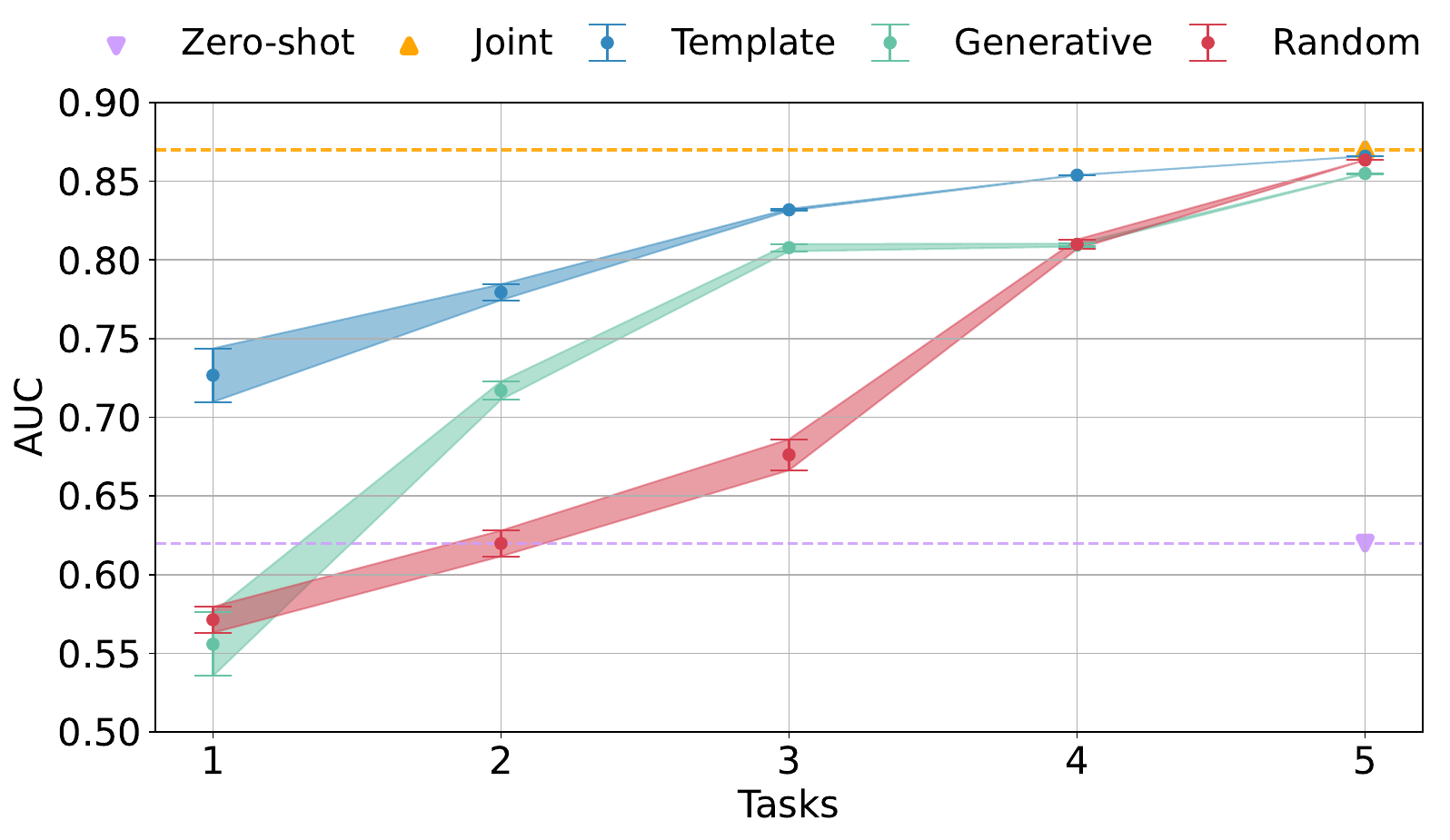}
        \caption{Label incremental}
        \label{fig:label-sub}
    \end{subfigure}
    \begin{subfigure}{0.33\textwidth}
        \centering
        \includegraphics[width=\linewidth]{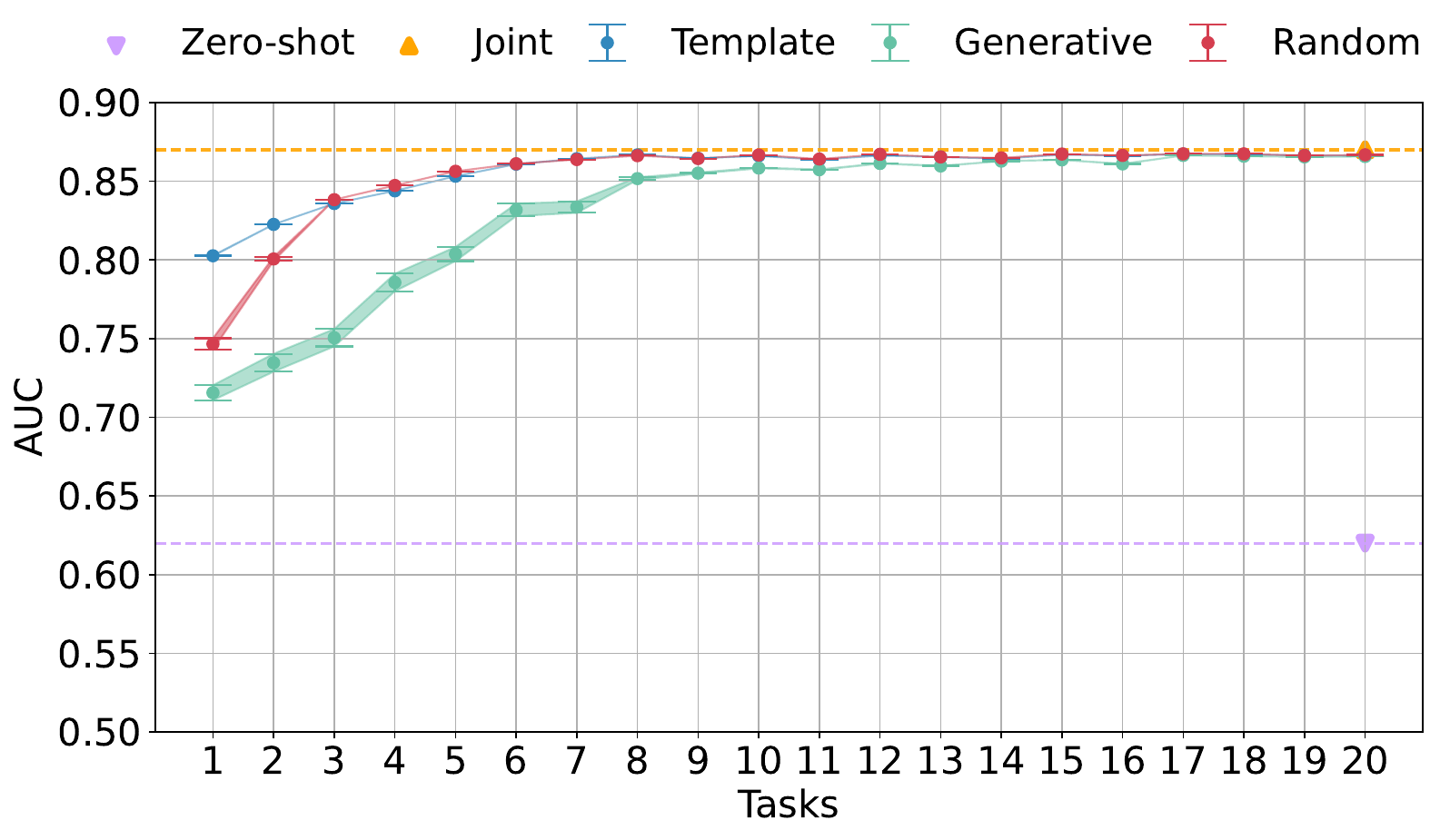}  
        \caption{Data incremental}
        \label{fig:data-sub}
    \end{subfigure}
    \captionsetup{skip=1pt}
    \caption{AUC performance by task for the three incremental scenarios (\textit{double-adaptor} shown). \vspace{-0.2in}}
    \label{fig:classlabdata}
\end{figure}
\begin{table}
    \centering
    \tiny
    \setlength{\tabcolsep}{1.6pt}
    \renewcommand{\arraystretch}{1.2} 
     \caption{Comparison of adaptor architectures and prompts in all scenarios.}

    \begin{tabular}{l|ccc|ccc|ccc|ccc}
           \hline
        \textbf{Adapter} & \multicolumn{3}{c|}{\textbf{MLP only on Image Encoder}} & \multicolumn{3}{c|}{\textbf{MLP only on Text Encoder}} & \multicolumn{3}{c|}{\textbf{MLP Shared between Image \& Text}} & \multicolumn{3}{c}{\textbf{MLP on Image \& Text}} \\
        \hline
        \textbf{Prompts} & \textbf{Template} & \textbf{Generative} & \textbf{Random} & \textbf{Template} & \textbf{Generative} & \textbf{Random} & \textbf{Template} & \textbf{Generative} & \textbf{Random} & \textbf{Template} & \textbf{Generative} & \textbf{Random} \\
        \hline
        \textbf{Joint} & 0.84$\pm$2e-6 & 0.70$\pm$4e-3 & 0.67$\pm$3e-2 & 0.85$\pm$1e-7 & 0.84$\pm$3e-4 & 0.85$\pm$9e-8 & \textbf{0.87}$\pm$4e-7 & 0.86$\pm$3e-4 & 0.86$\pm$1e-6 & \textbf{0.87}$\pm$5e-7 & \textbf{0.87}$\pm$2e-5 & \textbf{0.87}$\pm$4e-7 \\
        \textbf{Class-inc} & 0.81$\pm$8e-4 & 0.65$\pm$2e-2 & 0.59$\pm$6e-3 & 0.83$\pm$4e-5 & 0.73$\pm$1e-2 & 0.64$\pm$2e-2 & \textbf{0.84}$\pm$4e-4 & 0.62$\pm$3e-2 & 0.71$\pm$2e-2 & \textbf{0.84}$\pm$2e-4 & 0.73$\pm$1e-2 & 0.57$\pm$2e-2 \\
        \textbf{Label-inc} & 0.84$\pm$4e-5 & 0.72$\pm$3e-3 & 0.74$\pm$6e-3 & 0.85$\pm$1e-6 & 0.84$\pm$2e-4 & 0.84$\pm$1e-5 & 0.86$\pm$1e-5 & 0.85$\pm$4e-4 & 0.85$\pm$5e-5 & \textbf{0.87}$\pm$2e-6 & 0.85$\pm$2e-4 & 0.86$\pm$1e-5 \\
        \textbf{Data-inc} & 0.84$\pm$2e-6 & 0.70$\pm$4e-3 & 0.67$\pm$3e-2 & 0.85$\pm$7e-8 & 0.84$\pm$4e-5 & 0.85$\pm$9e-8 & \textbf{0.87}$\pm$5e-7 & 0.86$\pm$1e-5 & 0.86$\pm$9e-7 & \textbf{0.87}$\pm$2e-7 & \textbf{0.87}$\pm$1e-6 & \textbf{0.87}$\pm$2e-7 \\
        \hline
    \end{tabular}
    \vspace{-0.2in}
    \label{tab:tabellone}
\end{table}

RE-tune with Template Prompts demonstrates remarkable class-incremental learning capabilities (Fig.~\ref{fig:class-sub}): the use of semantically rich prompts as knowledge transmitters evidently prevents catastrophic forgetting between tasks even in such a complex setting. In the label- and data-incremental settings, RE-tune reaches joint training performance even using Generative or Random Prompts (Figs.~\ref{fig:label-sub} and \ref{fig:data-sub}). RE-tune in fact exploits the ability of the Image Encoder to differentiate between X-rays of diseased and non-diseased subjects based solely on visual features, and takes advantage of text prompt to create separate attractors, demonstrating that Biomedical VLMs are natural continual learners. Additionally, in the data-incremental scenario, RE-tune is data efficient, achieving upper bound performance using only 40\% of the dataset (Fig.~\ref{fig:data-sub}). Ablations on prompts and architectures validate the aforementioned phenomena (Tab.~\ref{tab:tabellone}). In Sec.~\ref{sec:2} we proposed to extend the pre-trained architecture with two adaptors, which is the most effective configuration. MLP on both Image and Text Encoders with Template Prompts yields the best performance in all evaluated scenarios.

Despite its robustness and efficiency, RE-tune is not without limitations. Our experiments show that the quality of textual prompts plays a pivotal role in performance. We experimented with automated prompt generation, but found that the best-performing prompts are still manually engineered ones. Also, while the model shows consistent performance across different adaptor variants, some scenarios exhibit a higher degree of noise (see Tab.~\ref{tab:tabellone}), indicating opportunities to further improve generalization. The use of text prompts as a knowledge transmission mechanism is an exciting direction for future research, with the potential to transform not just the medical imaging field but any domain where the ability to learn continuously and adaptively in privacy-preserving ways is of paramount importance. We believe this will help facilitate the adoption in real-world healthcare scenarios. 

\minisection{Potential Negative Societal Impacts.}
\label{sec:neg}
While incremental learning for chest X-ray diagnosis using approaches like RE-tune offers numerous benefits, it also raises certain potential problems and concerns that need to be carefully addressed to ensure its responsible and ethical deployment in healthcare settings:
\begin{itemize}
    \item \textbf{Bias and Fairness:} Incremental learning models are highly dependent on the data they are exposed to during training. If the initial dataset used for training is biased in terms of demographics, disease prevalence, or image quality, the model may inherit and perpetuate these biases. This could lead to disparities in diagnosis and treatment for different patient populations, exacerbating existing healthcare inequalities.
    \item \textbf{Data Privacy:} Privacy is a critical concern in healthcare, and incremental learning approaches must be designed with strong data privacy safeguards. While RE-tune is exemplar-free and privacy-preserving in terms of not storing patient-specific data, it still relies on access pre-deidentified medical image and medical reports, which can still contain sensitive patient information. Ensuring the secure handling of such data and implementing robust data anonymization techniques is essential to protect patient privacy.
    \item \textbf{Transparency and Explainability:} Incremental learning models, particularly those based on complex architectures like VLMs, can be difficult to interpret. Understanding how the model arrives at a particular diagnosis or decision is crucial for gaining the trust of healthcare practitioners and patients. Ensuring transparency and explainability in the decision-making process is an ongoing challenge in the field of AI in medicine.
\end{itemize}


\medskip
\bibliography{biblio}

\end{document}